\pdfoutput=1

\documentclass{article}
\usepackage{ijcai18}

\usepackage{times}
\usepackage{xcolor}
\usepackage{soul}
\usepackage[utf8]{inputenc}
\usepackage[small]{caption}

\usepackage{color}
\usepackage{amsmath,amssymb}
\usepackage{listings}
\usepackage{pgfplots}
\usepgfplotslibrary{units}
\pgfplotsset{compat=newest}

\usepackage{tikz}
\usetikzlibrary{shapes, positioning, patterns, plotmarks}
\usetikzlibrary{fit, calc, arrows, decorations.markings}
\usetikzlibrary{arrows,automata,backgrounds,fit}
\usepackage{subcaption}
\usepackage[normalem]{ulem}
\usepackage{multirow}
\usepackage{url}

\newcommand\Tstrut{}

\definecolor{mygray}{rgb}{0.5,0.5,0.5}
\lstdefinestyle{shell}{
	basicstyle=\small\ttfamily,
	numbers=none,
	columns=fullflexible
}

\usepackage{calc}
\usepackage{graphicx}
\usepackage{xcolor}
\newlength{\DepthReference}
\newlength{\HeightReference}
\newlength{\Width}%
\newcommand{\MyColorBox}[2][yellow]%
{%
	\settowidth{\Width}{#2}%
	\colorbox{#1}%
	{%
		\raisebox{-\DepthReference}%
		{%
			\parbox[b][\HeightReference+\DepthReference][c]{\Width}{\centering#2}%
		}%
	}%
}

\newcommand{\code}[1]{\texttt{#1}}

\title{Deep Reinforcement Learning for Programming Language Correction}

\author{
	Rahul Gupta, 
	Aditya Kanade, 
	Shirish Shevade
	\\ 
	Department of Computer Science and Automation \\
    Indian Institute of Science \\
    Bangalore, KA 560012, India \\
	\{rahulg,
	kanade,
	shirish\}@iisc.ac.in
}

\begin{document}
\maketitle

\begin{abstract}
Novice programmers often struggle with the formal syntax of
programming languages.
To assist them, we design a novel programming language correction
framework amenable to reinforcement learning.
The framework allows an agent to mimic human actions for text
navigation and editing.
We demonstrate that the agent can be trained through self-exploration directly from the raw input,
that is, program text itself, without any knowledge of the formal
syntax of the programming language.
We leverage expert demonstrations for one tenth of the training data to accelerate training.
The proposed technique is evaluated on $6975$ erroneous C programs with
typographic errors, written by students during an introductory programming
course.
Our technique fixes $14\%$ more programs and $29\%$ more compiler error
messages relative to those fixed by a state-of-the-art tool, DeepFix,
which uses a fully supervised neural machine translation approach.
\end{abstract}

\section{Introduction}
Programming languages prescribe formal rules of syntactic
validity of program text. A program text that does not adhere
to those rules cannot be compiled and executed. The inability
to write syntactically correct programs therefore becomes
a stumbling block for novice programmers. In large-scale online
programming courses, getting personalized feedback from instructors
or TAs is infeasible. On the other hand, compiler error
messages are often difficult to understand and do not localize
the errors accurately~\cite{Traver:2010:CEM:1863617.1945532}.
Therefore, our aim in this work is
to develop a technique to assist novice programmers by
automatically correcting common syntactic errors in programs.

We approach this problem through reinforcement learning (RL).
When faced with an error, a programmer navigates through
the program text to arrive at the location of error and then performs an edit 
to fix the error. We propose a novel {programming language correction} framework, in 
which an agent can mimic these actions. While the agent can access and modify the
program text, the compiler which checks syntactic validity of
the program text is a black-box for the agent. As noted above, compilers
usually do not pinpoint error locations precisely. Hence, we
do not rely on a compiler to aid in error localization or correction.
We instead design a reward function using only the \emph{number} of
error messages generated by the compiler. The goal of the agent 
is to perform edits necessary for successful compilation of the program.

The challenge in this framework is to learn
a control policy for the agent from program text without
the agent having any knowledge of the formal syntax of the programming language.
Through deep reinforcement learning, agents have been trained to 
play visual and text-based games at expert 
levels~\cite{mnih2015human,narasimhan2015language,wu2016training}. Interestingly, these
techniques learn directly from raw inputs such as pixels and text.
In this work, for the first time, we show that
this is possible even in the space of computer programs.

To see our framework in action, consider an example C program shown in Figure~\ref{fig:RLAssist-illustraion} which implements a tax calculation algorithm.
It has two syntactic errors: (1) incorrect use of semicolon within the \code{scanf} in line~\ref{line:error1} and (2) a missing closing brace at the end of line~\ref{line:error2}.
This program is given to a \emph{trained} agent which has not seen it during training.
The program is presented in a tokenized form to the agent and
the cursor position of the agent is initialized to the first token in the program.
The navigation actions of the agent over the program text are shown by arrows.
As shown by the sequence of actions taken by the agent in Figure~\ref{fig:RLAssist-illustraion}, the agent correctly localizes and fixes both the errors.
First, the agent navigates to the error location at line~\ref{line:error1} and replaces the incorrect semicolon
with a comma (marked by {$e_1$}).
Next, it navigates to the error location at line~\ref{line:error2} and inserts the missing closing brace (marked by {$e_2$}).
After these edits, the program compiles successfully and the agent stops.
In comparison, a brute-force search will have to enumerate all possible
edited versions of the program with up to two simultaneous edits and compile each of them
to identify the right edits. This will be far more expensive than using
the agent.

A video demonstration of this example is available at:
\begin{center}
\url{https://youtu.be/S4D6MR728zY}
\end{center}

\begin{figure}[t]
  \centering
  \begin{tikzpicture}[
      background rectangle/.style={fill=lightgray!35}, show background rectangle,
      every node/.style={font=\small, inner sep=0pt}]
	\newlength{\linesep}
	\setlength{\linesep}{5mm}
	\newlength{\tokensep}
	\setlength{\tokensep}{0.7mm}
	\tikzstyle{crossedge} = [->]

\node (one) at (0,0) {\begin{lstlisting}[firstnumber=1]
#include<stdio.h>
\end{lstlisting}};

\node [left=2mm of one.west] (dummy) {};

\node [below=\linesep of one.west,anchor=west] (two) {\begin{lstlisting}[firstnumber=2]
int main()(
\end{lstlisting}};
\node [xshift=2.5mm, below=\linesep of two.west,anchor=west] (inthree) {};
\node [below=\linesep of two.west,anchor=west] (three) {\begin{lstlisting}[firstnumber=3]
	float ti, tax;
\end{lstlisting}};
\node [xshift=2.5mm, below=\linesep of three.west,anchor=west] (infour) {};
\node [below=\linesep of three.west,anchor=west] (four) {\begin{lstlisting}[firstnumber=4, escapechar=!]
	scanf ( "%f" !\colorbox{red!50}{;}! &ti); !\label{line:error1}!
\end{lstlisting}};
\node [xshift=-8.0mm, below=\tokensep of three] (41s) {};
\node [xshift=-5.0mm, below=\tokensep of three] (41e) {};
\node [xshift=-1 mm, below=\tokensep of three] (42s) {};
\node [xshift=2 mm, below=\tokensep of three] (42e) {};
\node [xshift=4.8 mm, below=\tokensep of three] (43s) {};
\node [xshift=7.8 mm, below=\tokensep of three] (43e) {};
\node [circle, draw=black, scale=1, xshift=15mm, yshift=2.5mm, below=0mm of three, very thick] (44) {{{$e_1$}}};
\node [xshift=14 mm, below=5.4mm of three] (44down) {};

\node [xshift=2.5mm, below=4.5mm of four.west,anchor=west] (infive) {};
\node [below=\linesep of four.west,anchor=west] (five) {\begin{lstlisting}[firstnumber=5]
	if(ti<200001){
\end{lstlisting}};
\node [xshift=5mm, below=\linesep of five.west,anchor=west] (insix) {};
\node [below=\linesep of five.west,anchor=west] (six) {\begin{lstlisting}[firstnumber=6]
		printf("ti=0");}
\end{lstlisting}};
\node [xshift=2.5mm, below=\linesep of six.west,anchor=west] (inseven) {};
\node [below=\linesep of six.west,anchor=west] (seven) {\begin{lstlisting}[firstnumber=7]
	else if(200000<ti && ti<500001){
\end{lstlisting}};
\node [xshift=5mm, below=\linesep of seven.west,anchor=west] (ineight) {};
\node [below=\linesep of seven.west,anchor=west] (eight) {\begin{lstlisting}[firstnumber=8]
		tax=0.1*(ti-200000);
\end{lstlisting}};
\node [xshift=5mm, below=\linesep of eight.west,anchor=west] (innine) {};
\node [below=\linesep of eight.west,anchor=west] (nine) {\begin{lstlisting}[firstnumber=9]
		printf("%.2f", tax);}
\end{lstlisting}};
\node [xshift=2.5mm, below=\linesep of nine.west,anchor=west] (inten) {};
\node [below=\linesep of nine.west,anchor=west] (ten) {\begin{lstlisting}[firstnumber=10]
	else if(500000<ti && ti<1000001){
\end{lstlisting}};
\node [xshift=5mm, below=\linesep of ten.west,anchor=west] (ineleven) {};
\node [below=\linesep of ten.west,anchor=west] (eleven) {\begin{lstlisting}[firstnumber=11]
		tax=30000+0.2*(ti-500000);
\end{lstlisting}};
\node [xshift=5mm, below=\linesep of eleven.west,anchor=west] (intwelve) {};
\node [below=\linesep of eleven.west,anchor=west] (twelve) {\begin{lstlisting}[firstnumber=12, escapechar=!]
		printf ( "%.2f" , tax ) ; !\colorbox{red!50}{\vphantom{;}}! !\label{line:error2}!
\end{lstlisting}};
\node [xshift=-15.5mm, below=\tokensep of eleven] (121) {};
\node [xshift=-12.5mm, below=\tokensep of eleven] (121e) {};
\node [xshift=-8mm, below=\tokensep of eleven] (122) {};
\node [xshift=-5mm, below=\tokensep of eleven] (122e) {};
\node [xshift=-0.5mm, below=\tokensep of eleven] (123) {};
\node [xshift=2.5mm, below=\tokensep of eleven] (123e) {};
\node [xshift=7mm, below=\tokensep of eleven] (124) {};
\node [xshift=10mm, below=\tokensep of eleven] (124e) {};
\node [xshift=12mm, below=\tokensep of eleven] (125) {};
\node [xshift=15mm, below=\tokensep of eleven] (125e) {};
\node [xshift=17mm, below=\tokensep of eleven] (126) {};
\node [xshift=20mm, below=\tokensep of eleven] (126e) {};
\node [xshift=20.5mm, below=\tokensep of eleven] (127) {};
\node [xshift=23.5mm, below=\tokensep of eleven] (127e) {};
\node [circle, draw=black, scale=1, xshift=26mm, yshift=2.5mm, below=0mm of eleven, very thick] (128) {{{$e_2$}}};

\node [below=\linesep of twelve.west,anchor=west] (thirteen) {\begin{lstlisting}[firstnumber=13]
	else if(ti>1000000){
\end{lstlisting}};
\node [below=\linesep of thirteen.west,anchor=west] (fourteen) {\begin{lstlisting}[firstnumber=14]
		tax=130000+0.3*(ti-1000000);
\end{lstlisting}};
\node [below=\linesep of fourteen.west,anchor=west] (fifteen) {\begin{lstlisting}[firstnumber=15]
		printf("%.2f", tax);}
\end{lstlisting}};
\node [below=\linesep of fifteen.west,anchor=west] (sixteen) {\begin{lstlisting}[firstnumber=16]
	return 0;}
\end{lstlisting}};

\draw [->, shorten <= 0.6mm, shorten >= 1mm, bend right, very thick] (one.west) to (two.west);
\draw [->, shorten <= 0.6mm, shorten >= 1mm, bend right, very thick] (two.west) to (inthree.west);
\draw [->, shorten <= 0.6mm, shorten >= 1mm, bend right, very thick] (inthree.west) to (infour.west);

\draw [->, very thick] (41s) to (41e);
\draw [->, very thick] (42s) to (42e);
\draw [->, very thick] (43s) to (43e);
\draw [->, shorten <= 1mm, shorten >= 1.5mm, out=190, in=30, very thick] (44down.west) to (infive.north);

\draw [->, shorten <= 0.6mm, shorten >= 1mm, bend right, very thick] (infive.west) to (insix.west);
\draw [->, shorten <= 0.6mm, shorten >= 1mm, bend right, very thick] (insix.west) to (inseven.west);
\draw [->, shorten <= 0.6mm, shorten >= 1mm, bend right, very thick] (inseven.west) to (ineight.west);
\draw [->, shorten <= 0.6mm, shorten >= 1mm, bend right, very thick] (ineight.west) to (innine.west);
\draw [->, shorten <= 0.6mm, shorten >= 1mm, bend right, very thick] (innine.west) to (inten.west);
\draw [->, shorten <= 0.6mm, shorten >= 1mm, bend right, very thick] (inten.west) to (ineleven.west);
\draw [->, shorten <= 0.6mm, shorten >= 1mm, bend right, very thick] (ineleven.west) to (intwelve.west);

\draw [->, very thick] (121) to (121e);
\draw [->, very thick] (122) to (122e);
\draw [->, very thick] (123) to (123e);
\draw [->, very thick] (124) to (124e);
\draw [->, very thick] (125) to (125e);
\draw [->, very thick] (126) to (126e);
\draw [->, very thick] (127) to (127e);

\end{tikzpicture}
\caption{An erroneous program and the sequence of actions taken by a trained agent to
  fix it: The error locations are highlighted in the red color. The arrows show
  how the agent navigates over the program text. The edit actions are marked
  by $e_1$ and $e_2$.
}
\label{fig:RLAssist-illustraion}
\end{figure}

We encode the program text, augmented with the position of the cursor,
using a long short-term memory (LSTM) network~\cite{hochreiter1997long}.
The agent is allowed a set of navigation and edit actions to fix the program.
It receives a small reward for every edit which fixes some error and the maximum reward is given for reaching the goal state, which is an error-free version of the program.
The control policy of the agent is learned using the asynchronous advantage actor-critic (A3C) algorithm~\cite{mnih2016asynchronous}.

Training an agent in this setting is non-trivial for two reasons.
First, as illustrated by the example above, the agent needs to both localize
the errors and make precise edits at the error locations to be able to fix a program.
A wrong edit only makes the program worse by introducing more errors and
consequently, makes the task even more difficult.
To overcome this, we configure the environment to reject such edits.
This significantly prunes the state space that an agent is allowed to explore
without sacrificing its ability to fix programs. 
Second, reinforcement learning tends to be slow for the tasks with large state spaces as the time required for gathering information by state exploration increases.
One way to mitigate this problem is to guide the agent with expert demonstrations~\cite{Argall-2009-17073}.
Motivated by this, we design a scheme to 
enable the agent to take advantage of expert demonstrations
illustrating the required actions for fixing the programs.
As explained later, for our task, these demonstrations are generated automatically without any human intervention.
We call our technique \emph{RLAssist}.

DeepFix~\cite{gupta2017deepfix} is a state-of-the-art tool for fixing common
programming errors. 
We compare RLAssist with DeepFix on the task of fixing
typographic errors in $6975$ C programs. These programs were written by students
during an introductory programming course and
span $93$ different programming problems~\cite{gupta2017deepfix,prutor}\footnote{\small{\url{https://www.cse.iitk.ac.in/users/karkare/prutor/prutor-deepfix-09-12-2017.db.gz}}}.
These programs use non-trivial constructs of 
the C language such as multiple procedures, conditionals, switch statements, nested loops,
multi-dimensional arrays and recursion.
DeepFix uses a fully supervised neural machine translation approach.

In contrast, RLAssist is a deep reinforcement learning based technique.
We demonstrate that RLAssist can be trained through self-exploration using erroneous programs only, while still matching the performance of DeepFix.
We further show that we can leverage expert demonstrations to accelerate the training.
In particular, we generate expert demonstrations for one tenth of the training dataset. The rest $ 90\% $ of the dataset is used without demonstrations.
RLAssist fixes $26.6\%$ programs from the test set completely and resolves $39.7\%$ error messages.
Relative to DeepFix, this is an improvement of $14\%$ and $29\%$ respectively.
Thus, RLAssist outperforms DeepFix using expert demonstrations for a fraction of training data.

The main contributions of this work are as follows:
\begin{enumerate}
\item We design a novel framework of programming language correction amenable to
  reinforcement learning.
\item We use A3C for programming language correction and accelerate training 
through expert demonstrations.
\item Our experiments show that our technique outperforms a state-of-the-art tool, DeepFix, using expert demonstrations for only one tenth of the training data.
Moreover, our technique also works without any demonstrations while still matching the performance of DeepFix.
 \item The implementation of RLAssist will be open sourced.
\end{enumerate}

\section{Related Work}
Natural language correction is a well researched problem.
Some of the earlier works in this area have focused on identifying and correcting specific types of grammatical errors such as misuse of verb forms, articles, and prepositions ~\cite{chodorow2007detection,han2006detecting,rozovskaya2010generating}.
The more recent works consider a broader range of error classes, often relying on language models, and machine translation~\cite{ng2014conll,rozovskaya2014illinois}.

Although natural languages and programming languages are similar to some extent,
the latter have procedural interpretation and richer structure.
Due to the huge popularity of programming oriented
massive open online courses (MOOCs),
recent years have seen increasing interest in programming language correction~\cite{gupta2017deepfix,parihar2017automatic}.
These works either rely on supervised learning techniques
or use hand written rules to fix common programming errors.
This paper proposes a reinforcement learning framework in which an agent
can learn programming language correction directly from raw program text through self-exploration.

Learning from demonstration~(LfD) approaches train an agent using expert demonstrations.
Behavioral cloning~\cite{DBLP:journals/jmlr/RossGB11} is one particular class of LfD techniques to make the agent mimic expert demonstrations using supervised learning.
We complement the self-exploration with expert demonstrations to accelerate the training.
Inverse reinforcement learning is another class of LfD techniques.
These use demonstrations to first infer a reward function which is then used to learn the policy~\cite{Abbeel:2004:ALV:1015330.1015430}.
These methods have been used for the tasks where there is no obvious reward function, e.g., autonomous driving~\cite{Abbeel:2004:ALV:1015330.1015430} and acrobatic helicopter maneuvers~\cite{abbeel2007application}.
For our task, the rewards are well defined and can be calculated easily.

\section{Background}

\subsection{Reinforcement Learning}
In reinforcement learning, an agent interacts with its environment over a number of discrete time steps.
At each time step $ t $, the environment presents a state $ s_{t} \in \mathcal{S} $ to the agent. 
In response, the agent selects an action $ a_{t} $ from the set of allowed actions $ \mathcal{A} $.
This selection is controlled by the agent's policy $ \pi(a|s) = Pr\{a_t = a | s_t = s\} $.
The action is passed on to the environment and its execution 
may modify the internal state of the environment. 
The agent then receives the updated state $ s_{t+1} $ and a scalar reward  $ r_{t} $. 
This interaction, which is also called an episode,  stops when the agent reaches a goal state.
The objective of the agent is to maximize the expected sum of discounted future rewards $ G_{t} $ from each state $ s_t $.
For an episode terminating at time step $ T $, $ G_{t} = \sum_{k=0}^{T-t-1}\gamma^{k} r_{t+k} $, where the discount rate $ \gamma \in (0, 1] $~\cite{sutton1998reinforcement}.

\subsection{Asynchronous Advantage Actor-Critic (A3C)}
One of the many ways to solve an RL problem is to use policy gradient methods.
These methods learn a parameterized policy $ \pi(a|s; \theta) $, where $ \theta $ represents the parameters of a function approximator, such as a neural network.
One example of policy gradient methods is the actor-critic methods, which learn both $ \pi(a|s; \theta) $, the `actor', and the value function $ V(s; w) $, the `critic'.
Here, $ V(s; w) $ defines the expected reward from state $ s $, with $ w $ being the parameters of a function approximator. 
The critic evaluates how advantageous is it to be in the new state reached by taking an action $ a_t $ sampled from the distribution given by $ \pi(s_t) $.
Based on this evaluation, the parameterized policy is updated using an appropriate optimization technique such as gradient ascent~\cite{sutton1998reinforcement}.

The A3C algorithm uses multiple asynchronous parallel actor-learner threads to update a shared model, stabilizing the learning process by reducing the correlation of an agent's experience.
It has also been observed to reduce training time by a factor that is roughly linear in the number of parallel actor-learners~\cite{mnih2016asynchronous}.
We use A3C in this work.

\section{Technical Details}

\subsection{A Framework for Programming Language Correction Tasks}
When faced with an error,
a programmer navigates the program text to arrive at the location of error and then performs an edit operation to fix the error.
In the presence of multiple errors, the programmer can repeat these steps.
We present a programming language (PL) correction framework in which an agent can mimic these actions.
We now describe the components of this framework and their instantiation for our task of
correcting syntactic errors in C programs.

\subsubsection{States}
A state is a pair $\langle string, cursor \rangle$, where $ string $ is the program text, and $ cursor \in \{ 1, \ldots, len(string) \} $, where $ len(string) $ denotes the number of tokens in the $ string $.
The environment also keeps track of the number of errors in $ string $.
These errors can either be determined from the ground truth whenever available or estimated using the error messages generated by a compiler upon compiling the $string$.
For compilation, we use the GNU C compiler.
	
We encode the state into a sequence of tokens as follows.
First, we convert the program $ string $ into a sequence of lexemes. 
The lexemes are of different types, such as keywords, operators, types, functions, literals, and variables.
In addition, we also retain line-breaks as lexemes to allow
two-dimensional navigation actions over the program text.
The $ cursor $ part of the state is represented by a special token, which is inserted in the sequence of lexemes right after the token whose index is held by $ cursor $.

Next, we build a shared vocabulary across all programs.
Except some common library functions such as \texttt{printf} and \texttt{scanf}, 
all other function and variable identifiers are mapped to a special token ID. 
Similarly, all the literals are mapped to special tokens according to their type, e.g., numbers to \texttt{NUM} and strings to \texttt{STR}.
All remaining tokens are included in the vocabulary without any modifications.
This mapping reduces the size of the vocabulary seen by the agent.

Note that this encoding is only required for feeding the state to the agent.
The actions predicted by the agent based on this encoding are executed on the original program $ string $ by the environment.
	
\subsubsection{Actions and Transitions}
The agent actions are divided into two categories, the first which update the $ cursor $ and the second which modify the $ string $.
We refer to the first category of actions as \emph{navigation actions} and the latter as \emph{edit actions}. 
The navigation actions allow an agent to navigate through the $ string $. These actions only change the $ cursor $ of a state and not the $ string $.
The edit actions on the other hand, are used for error correction.
They only modify the $ string $ and not the $ cursor $.
Wrong edit actions introduce more errors in the $ string $ rather than fixing them.
We configure the environment to reject all such edits to prune the state space from which fixing the program becomes even more difficult.
In addition, \emph{rejecting wrong edits prevents arbitrary changes to the program.}

For our task, we allow only two navigation actions, \emph{move\_right} and \emph{move\_down}.
These set the $ cursor $ to the next token on the right or to the first token of the next line respectively.
The \emph{move\_right}~(respectively, \emph{move\_down}) action has no effect if the $ cursor $ is already set to the last token of a line~(respectively, any token of the last line).
Note that the \emph{move\_down} action is possible because we retain the line-breaks in the state encoding.

Based on a study of common typographic errors that novice programmers
make, we design three types of edit actions.
The first is a parameterized \emph{insert\_token} action which inserts the parameter \emph{token} immediately before the $ cursor $ position.
The parameter can be any token from a fixed set of tokens which we call \emph{mutable tokens}.
The second is the \emph{delete} action, which deletes the token at the $ cursor $ position.
However, the token is deleted only if it is from the set of mutable tokens.
We restrict the set of mutable tokens to the following five types
of tokens: semicolon, parentheses, braces, period, and comma.
The third edit action is a parameterized \emph{replace token1 with token2} action which replaces \emph{token1} at the cursor position with \emph{token2}. We have the following four actions in this class: (1)~\emph{replace `\code{;}' with `\code{,}'},
(2)~\emph{replace `\code{,}' with `\code{;}'},
(3)~\emph{replace `\code{.}' with `\code{;}'}, and (4)~\emph{replace `\code{;)}' with `\code{);}'}.
Although atomic replacement actions can be substituted with a sequence of delete and insert actions, having them prevents the cases where the constituent delete and/or insert actions can be rejected by the environment.

\subsubsection{Episode, Termination, and Rewards} An episode starts with an erroneous program text as $ string $ and the $ curosr $ set to its first token. 
The goal state is reached when the edited program compiles successfully.
An agent is allowed $ max\_episode\_len $ number of discrete time steps to reach the goal state in an episode after which the episode is terminated. 
Also, the agent is allowed only one pass over the program in an episode, i.e., once the agent navigates past the last token of a program, the episode is terminated. 

In each step, the agent is penalized with a small $ step\_penalty $ in order to encourage it to learn to fix a program in the minimum number of steps.
Additionally, the agent is penalized with a relatively higher $ edit\_penalty $ for taking edit actions.
The reason is that the edit actions are somewhat costly as they need to be verified by invoking a compiler. Therefore, we discourage the agent to make edits unnecessarily.
The agent is given $maximum\_reward$ for reaching the goal state.
Also, a small $ intermediate\_reward $ is given for taking an edit action that fixes at least one error.

\subsection{Model}
We use the A3C algorithm in the PL correction task. 
Our model first uses a long short-term memory (LSTM)~\cite{hochreiter1997long} network for embedding the tokenized state into a real vector. 
The LSTM network maps each token $ x_i $ of an input sequence $ (x_1, \ldots, x_n) $ to a real vector $ y_i $.
The final state embedding is calculated by taking an element-wise mean over all the output vectors $ (y_1, \ldots, y_n) $ following~\cite{narasimhan2015language}.

Given this state embedding, we use two separate fully connected linear layers to produce the policy function $ \pi(a|s; \theta) $, and the value function $ V(s; w) $ outputs.
Finally, before updating the network parameters, the gradients are accumulated using the following rules:
\begin{align*}
d\theta {}\leftarrow{} & d\theta + \nabla_{\theta^\prime} \log \pi(a_t|s_t;\theta^{\prime})(R - V(s_t;w^{\prime}))\\
& + \beta \nabla_{\theta^\prime} H(\pi(s_t;\theta^{\prime}))\\
dw {}\leftarrow{} & dw + \nabla_{w^\prime} (R - V(s_t; w^{\prime}))^2
\end{align*}
where $R = \sum_{i=0}^{k-1}\gamma^{i} r_{t+i} + \gamma^{k} V(s_{t+k}; w^{\prime})$, $ H $ is the entropy, and $ \beta $ is a hyperparameter to control the weight of the entropy regularization term; $ \theta^{\prime} $ and $ w^{\prime} $ are the thread specific parameters corresponding to $ \theta $ and $ w $ respectively~\cite{mnih2016asynchronous}.

\begin{table*}[t]
	\centering
	\scalebox{0.93}{
		\begin{tabular}{@{}|@{\;}r@{\;}|@{\;}r@{\;}|@{\;}r@{\;}|@{\;}r@{\;}|@{\;}r@{\;}|@{\;}r@{\;}|@{\;}r@{\;}|@{\;}r@{\;}|@{}}
			\hline
			\multicolumn{2}{|c|@{\;}}{\textbf{Dataset statistics}} &
			\multicolumn{3}{c|@{\;}}{\textbf{DeepFix results}} &
			\multicolumn{3}{c|}{\textbf{RLAssist results}} \Tstrut\\
			\hline
			\multicolumn{1}{|c|@{\;}}{Erroneous} & \multicolumn{1}{c|@{\;}}{Error} & \multicolumn{1}{c|@{\;}}{Completely} & \multicolumn{1}{c|@{\;}}{Partially}  &
			\multicolumn{1}{c|@{\;}}{Error} & 
			\multicolumn{1}{c|@{\;}}{Completely} & \multicolumn{1}{c|@{\;}}{Partially} &
			\multicolumn{1}{c|}{Error} \Tstrut\\
			\multicolumn{1}{|c|@{\;}}{programs} & \multicolumn{1}{c|@{\;}}{msgs.} & \multicolumn{1}{c|@{\;}}{fixed} & \multicolumn{1}{c|@{\;}}{fixed} &
			\multicolumn{1}{c|@{\;}}{messages} & \multicolumn{1}{c|@{\;}}{fixed} & \multicolumn{1}{c|@{\;}}{fixed} & \multicolumn{1}{c|}{messages} \Tstrut\\
			&  & \multicolumn{1}{c|@{\;}}{programs} & \multicolumn{1}{c|@{\;}}{programs} & \multicolumn{1}{c|@{\;}}{resolved} & \multicolumn{1}{c|@{\;}}{programs} & \multicolumn{1}{c|@{\;}}{programs} & \multicolumn{1}{c|}{resolved} \Tstrut\\
			\hline
			$6975$ & $16766$ & $1625$ ($23.3\%$) & $1129$ ($16.2\%$) & $5156$ ($30.8\%$) & $1854$ ($26.6\%$) & $1426$ ($20.4\%$) & $6652$ ($39.7\%$) \Tstrut\\
			\hline
		\end{tabular}
	}
	\caption{Summary of the test dataset and performance comparison of DeepFix and RLAssist on it.}
	\label{tab:datasets}
\end{table*}

\pgfplotstableread[col sep=comma]{ge_01-sr_bin_0.csv}\RLA
\pgfplotstableread[col sep=comma]{ge_00-sr_bin_0.csv}\ACOne
\pgfplotstableread[col sep=comma]{ge_00_bin_0.csv}\ACTwo

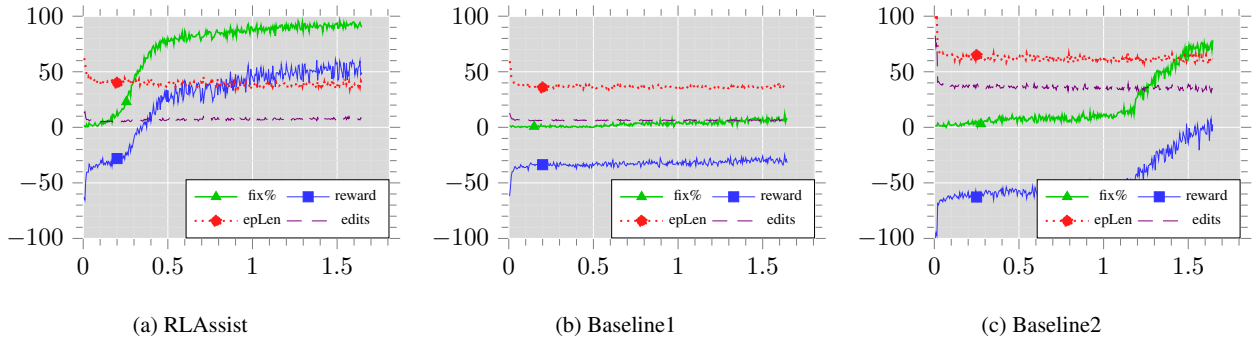
\begin{figure*}[t]
	\centering
	\scalebox{0.95}{
	\begin{subfigure}[t]{0.33\textwidth}
		\begin{tikzpicture}[anchor=base, baseline=(current bounding box.north)]
		\begin{axis}[
		axis line style={white},
		legend style={at={(1,0)}, anchor=south east, font=\tiny\selectfont, legend columns=2},
		scaled ticks=false,
		width=\textwidth,
		height=0.8\textwidth,
		grid=both,
		minor grid style={line width=.1pt, draw=black!12},
		major grid style={line width=.2pt,draw=white!35},
		xtick={0,500000,1000000,1500000},
		xticklabels={$ 0 $,$ 0.5 $,$ 1 $,$ 1.5 $},
		ymin=-100,ymax=100,
		ytick={-100,-50,0,50,100},
		ylabel near ticks,
		xtick align=outside,
		ytick align=outside,
		minor tick num=4,
		xmin=0,
		axis background/.style={fill=black!15},
		legend image post style={mark indices={}},
		]
		\addplot[semithick, color=black!20!green, mark=triangle*, mark indices={51}] table [x=episodes, y=fix_percent] \RLA;
		\addplot[color=blue!80, mark=square*, mark indices={40}] table [x=episodes, y=ep_rewards]  \RLA;
		\addplot[thick, color=red!90, mark=*, mark indices={40}, dash pattern=on \pgflinewidth off 2pt] table [x=episodes, y=ep_lens]  \RLA;
		\addplot[color=violet, dash pattern=on 6pt off 4pt] table [x=episodes, y=edits]  \RLA;
		\legend{fix\%, reward, epLen, edits}
		\end{axis}
		\end{tikzpicture}
		\caption{RLAssist}
		\label{fig:RLA}
	\end{subfigure}
	\begin{subfigure}[t]{0.33\textwidth}
		\begin{tikzpicture}[anchor=base, baseline=(current bounding box.north)]
		\begin{axis}[
		axis line style={white},
		legend style={at={(1,0)}, anchor=south east, font=\tiny\selectfont, legend columns=2},
		scaled ticks=false,
		width=\textwidth,
		height=0.8\textwidth,
		grid=both,
		minor grid style={line width=.1pt, draw=black!12},
		major grid style={line width=.2pt,draw=white!35},
		xtick={0,500000,1000000,1500000},
		xticklabels={$ 0 $,$ 0.5 $,$ 1 $,$ 1.5 $},
		ymin=-100,ymax=100,
		ytick={-100,-50,0,50,100},
		ylabel near ticks,
		xtick align=outside,
		ytick align=outside,
		minor tick num=4,
		xmin=0,
		axis background/.style={fill=black!15},
		legend image post style={mark indices={}},
		]
		\addplot[semithick, color=black!20!green, mark=triangle*, mark indices={30}] table [x=episodes, y=fix_percent] \ACOne;
		\addplot[color=blue!80, mark=square*, mark indices={40}] table [x=episodes, y=ep_rewards]  \ACOne;
		\addplot[thick, color=red!90, mark=*, mark indices={40}, dash pattern=on \pgflinewidth off 2pt] table [x=episodes, y=ep_lens]  \ACOne;
		\addplot[color=violet, dash pattern=on 6pt off 4pt] table [x=episodes, y=edits]  \ACOne;
		\legend{fix\%, reward, epLen, edits}
		\end{axis}
		\end{tikzpicture}
		\caption{Baseline1}
		\label{fig:baseline1}
	\end{subfigure}
	\begin{subfigure}[t]{0.33\textwidth}
		\begin{tikzpicture}[anchor=base, baseline=(current bounding box.north)]
		\begin{axis}[
		axis line style={white},
		legend style={at={(1,0)}, anchor=south east, font=\tiny\selectfont, legend columns=2},
		scaled ticks=false,
		width=\textwidth,
		height=0.8\textwidth,
		grid=both,
		minor grid style={line width=.1pt, draw=black!12},
		major grid style={line width=.2pt,draw=white!35},
		xtick={0,500000,1000000,1500000},
		xticklabels={$ 0 $,$ 0.5 $,$ 1 $,$ 1.5 $},
		ymin=-100,ymax=100,
		ytick={-100,-50,0,50,100},
		ylabel near ticks,
		xtick align=outside,
		ytick align=outside,
		minor tick num=4,
		xmin=0,
		axis background/.style={fill=black!15},
		legend image post style={mark indices={}},
		]
		\addplot[semithick, color=black!20!green, mark=triangle*, mark indices={55}] table [x=episodes, y=fix_percent] \ACTwo;
		\addplot[color=blue!80, mark=square*, mark indices={50}] table [x=episodes, y=ep_rewards]  \ACTwo;
		\addplot[thick, color=red!90, mark=*, mark indices={50}, dash pattern=on \pgflinewidth off 2pt] table [x=episodes, y=ep_lens]  \ACTwo;
		\addplot[color=violet, dash pattern=on 6pt off 4pt] table [x=episodes, y=edits] \ACTwo;
		\legend{fix\%, reward, epLen, edits}
		\end{axis}
		\end{tikzpicture}
		\caption{Baseline2}
		\label{fig:baseline2}
	\end{subfigure}
	}
	\caption{Training performance comparison of RLAssist with two baselines on one fold of the training dataset. None of the baselines uses expert demonstrations. Additionally, Baseline2 has $ edit\_penalty $ set to zero. The X-axis shows the number of training episodes in millions.
	\emph{epLen} and \emph{fix\%} stand for episode length and the percentage of error messages resolved, respectively.}
	\label{fig:ED-comparison}
\end{figure*}

\subsubsection{Expert Demonstrations}
As we show later in the experiments section, RLAssist can be trained without using any demonstrations, but requires longer training time.
The reason is that in the A3C algorithm, in each episode, an agent starts with a random state $ s $ and then uses its policy function to interact with the environment.
As the policy network is initialized randomly, this interaction is governed with random exploration at the beginning of the training.
With only random exploration, the agent finds it difficult to reach the goal state and wins rewards infrequently. Consequently, the training slows down.

We use expert demonstrations to accelerate training. 
An expert demonstration is a sequence of actions leading to the goal state of an episode.
Given a pair $ (p, p^{\prime}) $ where $ p $ is an incorrect program and $ p^{\prime} $ is its correct version, a demonstration is generated automatically as follows.
Starting from the first token in $ p^{\prime} $, each unmodified line (w.r.t.\ $ p $) is skipped through a \emph{move\_down} action.
At the first erroneous line, the cursor is moved rightwards through a \emph{move\_right} action until reaching an error location and the appropriate edit action is generated. This process is repeated until the last error in the program is resolved.

We configure an agent to use expert demonstrations as follows.
For the episodes for which a demonstration is available, the agent follows the predetermined sequence of actions provided, instead of sampling driven by the policy.
The updates to the policy network parameters are made as if the predetermined action was sampled.
For the rest of the episodes, the agent takes the actions sampled using the policy, following the standard A3C algorithm.
Note that the demonstrations are provided at the episode level and not at a finer granularity of transition level.
The reasoning is that the agent needs to take the right actions throughout the episode to reach the goal state and earn reward. 
If it takes intermittent guidance in an episode then it can still fail to reach the goal state.

\section{Experiments}

\subsection{Dataset}

We use the training and test datasets used in ~\cite{gupta2017deepfix} for the task of correcting typographic errors.
The programs in the datasets span $93$ programming problems in an introductory programming course and make use of non-trivial C language constructs.
The program lengths  range from $ 75 $ to $ 450 $ tokens.
In addition to the test set of actual student-written erroneous programs,
DeepFix has also been evaluated on some programs with seeded errors.
In this work, we only use the test set of student-written erroneous programs.
This test set contains $6975$ erroneous programs with
$16766$ compilation error messages together, as shown in Table~\ref{tab:datasets}.
The dataset is divided into five folds for cross validation.
Each fold holds aside about 1/5th of the $ 93 $ programming problems.
The student-written erroneous programs belonging to the held out problems
form the test set in that fold.
The training data associated with the remaining 4/5th programming problems is used for training.
Thus, a learning algorithm must learn the syntactic validity as per the language syntax so that it can generalize to unseen programming problems.
Each fold roughly contains about $160$K labeled training examples. 
For more details of the dataset, we refer the reader to~\cite{gupta2017deepfix}.

\subsection{Experiment Configuration and Training}
We implement our technique in Tensorflow~\cite{abadi2016tensorflow}, following an open source implementation\footnote{\small{\url{https://github.com/awjuliani/DeepRL-Agents/}}} of A3C.
We find a suitable configuration of the PL correction framework and the learning model for our task through experimentation.
In particular, the LSTM encoder in our model has two recurrent layers with $ 128 $ cells each. Our vocabulary has $ 91 $ tokens, which are embedded into $ 24 $-dimensional vectors.
We set the discounting factor $ \gamma = 0.99 $, the maximum number of exploration steps before a neural network parameter update is made $ t_{max} = 24 $, and entropy regularization factor $ \beta = 0.01 $. 
We use $ 32 $ parallel agents, and a learning rate of $ 0.0001 $ for optimizing our model using the ADAM optimizer~\cite{kingma2015adam}.
We also use gradient clipping to prevent the gradients from exploding~\cite{pascanu2013difficulty}.
We configure the PL correction framework for our task by setting $ max\_episode\_len = 100 $, $ step\_penalty = -0.005 $, $ edit\_penalty = -0.025 $, $ maximum\_reward = 1 $, and $ intermediate\_reward = 0.045 $.

In order to accelerate training, we use expert demonstrations for \emph{one tenth} of examples in the training dataset.
In our experiments, we observed that using more demonstrations did not result in significant speedup in training while reducing the demonstrations slowed it down.
With demonstrations, the training took about four days for $10$ epochs of training on an Intel(R) Xeon(R) E5-2630 v4 machine, clocked at 2.20GHz with 32GB of RAM.

\subsection{Evaluation}

In this section, we first discuss the training performance of RLAssist with expert demonstrations on one tenth of the training data. Next, we compare it with DeepFix on the test dataset described earlier. Later we compare RLAssist with two baselines which are trained without using any expert demonstrations. While the first baseline fails to learn at all, the second one is able to learn enough to match the performance of DeepFix, demonstrating that RLAssist can be trained using only erroneous programs and self-exploration.

\subsubsection{Training Performance of RLAssist}
In order to evaluate the training performance of RLAssist, we use the following metrics: (1) the percentage of error messages resolved, (2) the average episode length, (3) the average number of edit actions, and (4) the average reward obtained by the agent as the training progresses.
We report the training performance on one of the folds.
For ease of plotting, the average reward shown in the figures is scaled by a factor of $ 100 $.
Figure~\ref{fig:RLA} illustrates the training performance of RLAssist.

RLAssist learns to solve the task very well and is able to resolve about $ 90\% $ of the error messages after training for about a million episodes.
In the last epoch of training, RLAssist reaches the goal state for $ 85\% $ of the episodes.
Furthermore, it manages to resolve $ 56\% $ of the error messages for the programs corresponding to the remaining $ 15\% $ episodes.
At the same time, the average scaled reward also reaches the maximum of about $ 52 $ from $ -65 $, the reward obtained at the beginning of the training.
The maximum scaled reward is almost always less than $ 100 $ because of the penalty that the agent incurs for navigating to the error location.
The average length of an episode reaches about $ 38 $ consisting about $ 31 $ navigation and $ 7 $ edit actions.
For the $ 85\% $ of the programs that RLAssist fixes in the last epoch, the number of rejected edit actions per episode is less than $ 4 $.
This shows that RLAssist not only learns to fix a program but it also learns to do so by taking almost precise navigation and edit actions.

\subsubsection{Comparison with DeepFix}

\begin{table}[t]
	\centering
	\scalebox{0.93}{
		\begin{tabular}{@{}|@{\;}l@{\;}|@{\;}r@{\;}|@{\;}r@{\;}|@{\;}r@{\;}|@{}}
			\hline
			& \multicolumn{1}{c|@{\;}}{Completely} & \multicolumn{1}{c|@{\;}}{Partially}  & \multicolumn{1}{c|}{Completely or}\\
			& \multicolumn{1}{c|@{\;}}{fixed} & \multicolumn{1}{c|@{\;}}{fixed}  & \multicolumn{1}{c|}{partially fixed}\\
			& \multicolumn{1}{c|@{\;}}{programs} & \multicolumn{1}{c|@{\;}}{programs}  & \multicolumn{1}{c|}{programs}\\
			\hline 
			DeepFix alone & 199 & 179 & 248\\ \hline
			RLAssist alone & 425 & 476 & 771\\ \hline
		\end{tabular}
	}
	\caption{Comparison of DeepFix and RLAssist on the number of programs fixed exclusively by each technique.}
	\label{tab:comparison}
\end{table}

In Table~\ref{tab:datasets}, we show the comparison of RLAssist with DeepFix on the test dataset.
For this comparison, we use the number of error messages resolved, and completely and partially fixed programs; the same metrics as reported in~\cite{gupta2017deepfix}.
Further in Table~\ref{tab:comparison}, we also report the number of programs which are fixed exclusively by DeepFix or RLAssist.
For comparison, we take the most recent results available from the DeepFix webpage: {\small{\url{https://bitbucket.org/iiscseal/deepfix}}}.

As shown in Table~\ref{tab:datasets}, the test dataset has $ 16766 $ error messages from $ 6975 $ erroneous programs out of which DeepFix resolves $ 5156 $ error messages, fixing $ 1625 $ programs completely and $ 1129 $ partially.
RLAssist resolves $ 6652 $ error messages, fixing $ 1854 $ programs completely and $ 1426 $ partially.
Thus RLAssist outperforms DeepFix by a relative margin of $ 29\% $ and $ 14\% $ in terms of error messages resolved and completely fixed programs, respectively.
Relative to DeepFix, the percentage of programs fixed partially by RLAssist is $26\%$ higher.
At test time, both RLAssist and DeepFix take less than a second to fix a program.

One reason for better performance of RLAssist is that it works at a finer token-level granularity compared to the coarser line-level granularity of DeepFix.
RLAssist can edit an incorrect line in place, whereas DeepFix has to
produce a complete replacement of the erroneous line. This requires it to
copy the correct token subsequences from the input while simultaneously rectifying
the erroneous tokens.
Another reason is that DeepFix halts when it cannot fix a line, i.e., if it fails to fix an erroneous line, it cannot fix the subsequent erroneous lines.
This limitation arises because of the iterative nature of DeepFix.
If a fix suggested by DeepFix is accepted by the compiler, the fix is applied and the updated program is shown to
DeepFix to identify the next fix.
However, if it is rejected, the iterative procedure stops.
RLAssist, on the other hand, does not have such a limitation.
If the action taken by the agent is rejected by the environment, it takes the next highest value action and continues to attempt other fixes.

\subsubsection{Comparison with Baselines}

We now discuss two baselines which do \emph{not} use expert demonstrations for training.
Baseline1 is identical to RLAssist except that it does not use demonstrations.
Figure~\ref{fig:baseline1} shows the training performance of Baseline1. It shows that the training fails to make any progress even after $ 1.5 $ million episodes.
As the edit actions are penalized more than the navigation actions, the agent tries only about $ 5 $ edits per episode during exploration, causing the training to fail.

Considering this, in Baseline2, we set the $ edit\_penalty $ to zero.
As seen in Figure~\ref{fig:baseline2}, this baseline
starts showing signs of learning after a million episodes.
After about $ 1.5 $ million episodes (about $10$ epochs), it learns to resolve nearly $ 70\% $ of the error messages.
With additional $ 10 $ epochs~(not shown in Figure~\ref{fig:baseline2}), it resolves $ 76\% $ of the error messages.
Further, it matches the performance of DeepFix on the test dataset.
In fact, it slightly outperforms DeepFix.
This shows that RLAssist can also be trained entirely through self-exploration without using any expert demonstrations.

\subsubsection{Embedding Visualization}

\begin{figure}[t]
	\centering
	\begin{tikzpicture}[anchor=base, baseline=(current bounding box.north)]
	\begin{axis}[%
	scatter/classes={%
		state1={mark=diamond,purple!85}, state2={mark=square,orange!95}, state3={mark=o,blue}
	},
	width=0.65\textwidth,
	height=0.35\textwidth,
	axis lines=middle, axis on top,
	view={-25}{-25},
	xtick=\empty,		
	ytick=\empty,
	ztick=\empty,
	]
	\addplot3[
	scatter,
	only marks,%
	scatter src=explicit symbolic,
	mark size=1pt,
	]%
	table[meta=label, x={x}, y={y}, z={z}, col sep=comma] {embedding.csv};
	\end{axis}
	\end{tikzpicture}
	\caption{PCA projection of embeddings of $ 350 $ programs in three different states.
		The first, the second, and the third states are shown by diamonds, squares, and circles, respectively.}
	\label{fig:embeddings}
\end{figure}
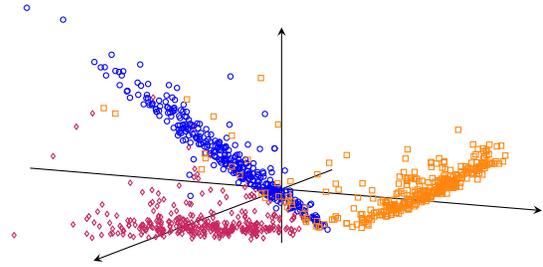

We select $ 350 $ test programs containing only one error per program.
Next, we get embeddings corresponding to the following three states of each of these programs: (1) when the cursor is set to the first token of the line preceding the erroneous line,
(2) when the cursor is set to the first token of the erroneous line,
and (3) when the cursor is set to the error location and the program has been fixed.
Figure~\ref{fig:embeddings} shows the first three principal components of these embeddings. It can be seen that the three states form three distinct clusters with almost no overlap. This shows that RLAssist's encoder learns to capture not only the syntactic validity of the programs but also the location of the errors in them.
This in turn helps the agent select appropriate actions, e.g., \emph{move\_down} in the first
state and \emph{move\_right} in the second state.

\section{Conclusions and Future Work}
We address the problem of programming language correction and present a novel deep reinforcement learning based solution, called RLAssist, for it.
We compare RLAssist with a state-of-the-art tool, DeepFix, on the task
of correcting typographic errors in $ 6975 $ student-written erroneous C programs.
Our experiments show that RLAssist outperforms DeepFix.
Moreover, we show that RLAssist can be trained without any expert demonstrations while still matching performance of DeepFix.

RLAssist is programming language agnostic and has been evaluated on C programs. 
In future, we will experiment with other programming languages as well.
We plan to extend RLAssist to target more classes of errors,
and devise RL algorithms that can learn and exploit deeper
syntactic and semantic properties of programs.

\section*{Acknowledgments}
We thank Sonata Software Ltd. for partially funding this work.

\bibliographystyle{named}
\bibliography{bibDB}

\begin{thebibliography}{}

\bibitem[\protect\citeauthoryear{Abadi \bgroup \em et al.\egroup
  }{2016}]{abadi2016tensorflow}
Mart{\'\i}n Abadi, Paul Barham, Jianmin Chen, Zhifeng Chen, Andy Davis, Jeffrey
  Dean, Matthieu Devin, Sanjay Ghemawat, Geoffrey Irving, Michael Isard, et~al.
\newblock Tensorflow: A system for large-scale machine learning.
\newblock In {\em Proceedings of the 12th USENIX Symposium on Operating Systems
  Design and Implementation}, volume~16, pages 265--283, 2016.

\bibitem[\protect\citeauthoryear{Abbeel and
  Ng}{2004}]{Abbeel:2004:ALV:1015330.1015430}
Pieter Abbeel and Andrew~Y. Ng.
\newblock Apprenticeship learning via inverse reinforcement learning.
\newblock In {\em Proceedings of the 21st International Conference on Machine
  Learning}, page~1, 2004.

\bibitem[\protect\citeauthoryear{Abbeel \bgroup \em et al.\egroup
  }{2007}]{abbeel2007application}
Pieter Abbeel, Adam Coates, Morgan Quigley, and Andrew~Y Ng.
\newblock An application of reinforcement learning to aerobatic helicopter
  flight.
\newblock In {\em Advances in neural information processing systems}, pages
  1--8, 2007.

\bibitem[\protect\citeauthoryear{Argall \bgroup \em et al.\egroup
  }{2009}]{Argall-2009-17073}
Brenna Argall, Sonia Chernova, Manuela Veloso, and Brett Browning.
\newblock A survey of robot learning from demonstration.
\newblock {\em Robotics and Autonomous Systems}, 67:469--483, 2009.

\bibitem[\protect\citeauthoryear{Chodorow \bgroup \em et al.\egroup
  }{2007}]{chodorow2007detection}
Martin Chodorow, Joel~R Tetreault, and Na-Rae Han.
\newblock Detection of grammatical errors involving prepositions.
\newblock In {\em Proceedings of the 4th ACL-SIGSEM workshop on prepositions},
  pages 25--30, 2007.

\bibitem[\protect\citeauthoryear{Das \bgroup \em et al.\egroup }{2017}]{prutor}
Rajdeep Das, Umair~Z. Ahmed, Amey Karkare, and Sumit Gulwani.
\newblock Prutor: A system for tutoring cs1 and collecting student programs for
  analysis, 2017.
\newblock \url{https://www.cse.iitk.ac.in/users/karkare/prutor/}.

\bibitem[\protect\citeauthoryear{Gupta \bgroup \em et al.\egroup
  }{2017}]{gupta2017deepfix}
Rahul Gupta, Soham Pal, Aditya Kanade, and Shirish Shevade.
\newblock Deep{F}ix: Fixing common c language errors by deep learning.
\newblock In {\em Proceedings of the 31st AAAI Conference on Artificial
  Intelligence}, pages 1345--1351, 2017.

\bibitem[\protect\citeauthoryear{Han \bgroup \em et al.\egroup
  }{2006}]{han2006detecting}
Na-Rae Han, Martin Chodorow, and Claudia Leacock.
\newblock Detecting errors in {E}nglish article usage by non-native speakers.
\newblock {\em Natural Language Engineering}, 12(2):115--129, 2006.

\bibitem[\protect\citeauthoryear{Hochreiter and
  Schmidhuber}{1997}]{hochreiter1997long}
Sepp Hochreiter and J{\"u}rgen Schmidhuber.
\newblock Long short-term memory.
\newblock {\em Neural computation}, 9(8):1735--1780, 1997.

\bibitem[\protect\citeauthoryear{Kingma and Ba}{2015}]{kingma2015adam}
Diederik Kingma and Jimmy Ba.
\newblock Adam: A method for stochastic optimization.
\newblock In {\em The 3rd International Conference on Learning
  Representations}, 2015.

\bibitem[\protect\citeauthoryear{Mnih \bgroup \em et al.\egroup
  }{2015}]{mnih2015human}
Volodymyr Mnih, Koray Kavukcuoglu, David Silver, Andrei~A. Rusu, Joel Veness,
  Marc~G. Bellemare, Alex Graves, Martin Riedmiller, Andreas~K. Fidjeland,
  Georg Ostrovski, Stig Petersen, Charles Beattie, Amir Sadik, Ioannis
  Antonoglou, Helen King, Dharshan Kumaran, Daan Wierstra, Shane Legg, and
  Demis Hassabis.
\newblock Human-level control through deep reinforcement learning.
\newblock {\em Nature}, 518(7540):529--533, 2015.

\bibitem[\protect\citeauthoryear{Mnih \bgroup \em et al.\egroup
  }{2016}]{mnih2016asynchronous}
Volodymyr Mnih, Adria~Puigdomenech Badia, Mehdi Mirza, Alex Graves, Timothy
  Lillicrap, Tim Harley, David Silver, and Koray Kavukcuoglu.
\newblock Asynchronous methods for deep reinforcement learning.
\newblock In {\em Proceedings of the 33rd International Conference on Machine
  Learning}, pages 1928--1937, 2016.

\bibitem[\protect\citeauthoryear{Narasimhan \bgroup \em et al.\egroup
  }{2015}]{narasimhan2015language}
Karthik Narasimhan, Tejas~D. Kulkarni, and Regina Barzilay.
\newblock Language understanding for text based games using deep reinforcement
  learning.
\newblock In {\em Proceedings of the Conference on Empirical Methods in Natural
  Language Processing}, pages 1--11, 2015.

\bibitem[\protect\citeauthoryear{Ng \bgroup \em et al.\egroup
  }{2014}]{ng2014conll}
Hwee~Tou Ng, Siew~Mei Wu, Ted Briscoe, Christian Hadiwinoto, Raymond~Hendy
  Susanto, and Christopher Bryant.
\newblock The {CoNLL}-2014 shared task on grammatical error correction.
\newblock In {\em CoNLL Shared Task}, pages 1--14, 2014.

\bibitem[\protect\citeauthoryear{Parihar \bgroup \em et al.\egroup
  }{2017}]{parihar2017automatic}
Sagar Parihar, Ziyaan Dadachanji, Praveen~Kumar Singh, Rajdeep Das, Amey
  Karkare, and Arnab Bhattacharya.
\newblock Automatic grading and feedback using program repair for introductory
  programming courses.
\newblock In {\em Proceedings of the ACM Conference on Innovation and
  Technology in Computer Science Education}, pages 92--97, 2017.

\bibitem[\protect\citeauthoryear{Pascanu \bgroup \em et al.\egroup
  }{2013}]{pascanu2013difficulty}
Razvan Pascanu, Tomas Mikolov, and Yoshua Bengio.
\newblock On the difficulty of training recurrent neural networks.
\newblock In {\em International Conference on Machine Learning}, pages
  1310--1318, 2013.

\bibitem[\protect\citeauthoryear{Ross \bgroup \em et al.\egroup
  }{2011}]{DBLP:journals/jmlr/RossGB11}
St{\'{e}}phane Ross, Geoffrey~J. Gordon, and Drew Bagnell.
\newblock A reduction of imitation learning and structured prediction to
  no-regret online learning.
\newblock In {\em Proceedings of the 14th International Conference on
  Artificial Intelligence and Statistics}, pages 627--635, 2011.

\bibitem[\protect\citeauthoryear{Rozovskaya and
  Roth}{2010}]{rozovskaya2010generating}
Alla Rozovskaya and Dan Roth.
\newblock Generating confusion sets for context-sensitive error correction.
\newblock In {\em Proceedings of the conference on empirical methods in natural
  language processing}, pages 961--970, 2010.

\bibitem[\protect\citeauthoryear{Rozovskaya \bgroup \em et al.\egroup
  }{2014}]{rozovskaya2014illinois}
Alla Rozovskaya, Kai-Wei Chang, Mark Sammons, Dan Roth, and Nizar Habash.
\newblock The illinois-columbia system in the {CoNLL}-2014 shared task.
\newblock In {\em CoNLL Shared Task}, pages 34--42, 2014.

\bibitem[\protect\citeauthoryear{Sutton and
  Barto}{1998}]{sutton1998reinforcement}
Richard~S Sutton and Andrew~G Barto.
\newblock {\em Reinforcement learning: An introduction}, volume~1.
\newblock MIT press Cambridge, 1998.

\bibitem[\protect\citeauthoryear{Traver}{2010}]{Traver:2010:CEM:1863617.1945532}
V.~Javier Traver.
\newblock On compiler error messages: What they say and what they mean.
\newblock {\em Advances in Human-Computer Interaction}, 2010:3:1--3:26, 2010.

\bibitem[\protect\citeauthoryear{Wu and Tian}{2017}]{wu2016training}
Yuxin Wu and Yuandong Tian.
\newblock Training agent for first-person shooter game with actor-critic
  curriculum learning.
\newblock In {\em The 5th International Conference on Learning
  Representations}, 2017.

\end{thebibliography}

\end{document}